# Strengthening the Training of Convolutional Neural Networks By Using Walsh Matrix.

Tamer Ölmez and Zümray Dokur

Istanbul Technical University,
Department of Electronics and Communication Engineering, Istanbul, Turkey
corresponding author: olmezt@itu.edu.tr

**ABSTRACT:** DNN structures are continuously developing and achieving high performances in classification problems. Also, it is observed that success rates obtained with DNNs are higher than those obtained with traditional neural networks. In addition, one of the advantages of DNNs is that there is no need to spend an extra effort to determine the features; the CNN automatically extracts the features from the dataset during the training. Besides their benefits, the DNNs have the following three major drawbacks among the others: (i) Researchers have struggled with over-fitting and under-fitting issues in the training of DNNs, (ii) determination of even a coarse structure for the DNN may take days, and (iii) most of the time, the proposed network structure is too large to be too bulky to be used in real time applications. We have modified the training and structure of DNN to increase the classification performance, to decrease the number of nodes in the structure, and to be used with less number of hyper parameters. A minimum distance network (MDN) following the last layer of the convolutional neural network (CNN) is used as the classifier instead of a fully connected neural network (FCNN). In order to strengthen the training of the CNN, we suggest employing Walsh function. We tested the performances of the proposed DNN (named as DivFE) on the classification of ECG, EEG, heart sound, detection pneumonia in X-ray chest images, detection of BGA solder defects, and patterns of benchmark datasets (MNIST, IRIS, CIFAR10 and CIFAR20). In different areas, it has been observed that a higher classification performance was obtained by using the DivFE with less number of nodes.

# 1. Introduction

From recent studies it is observed that convolutional neural networks are proved to be extremely successful in classification problems. Especially, accurate and fast classification of biological signals/images (EEG, ECG, heart sound and x-ray chest images etc.) is a crucial step in the implementation of real-time arrhythmia diagnosis systems. High classification performances are obtained by using the developed DNN structures. However, it is observed that the proposed network structure is too large to be too bulky to be used in real time applications. In this study, the use of the DivFE in different fields will be emphasized by demonstrating examples of our previous study.

It is observed that traditional neural network [1-7] and DNNs [8-16] are becoming popular in the classification of ECG signals. In the studies [8-16], high classification performances are obtained by using different deep neural network structures which generally consist of convolution neural network and fully connected neural networks. However, the extremely large size of the DNNs is still a major obstacle for real-time applications. For some patients suffering from cardiovascular problems, doctors may ask for the continuous recording of the patient's ECG in order to evaluate his/her heart activity on a long-term basis, and also to detect any cardiac symptom that does not show up during the ECG recording at a medical center, but may occur within a short time interval during the patient's normal daily routine. For these reasons, Holter monitors or some other similar systems, which are portable and wearable medical devices that measure and analyze the everyday activity of the heart, are being carried by selected patients. With such portable systems, if the doctor has sufficiently more information about the condition of the heart, it will be easier for him/her to deal with a vital problem.

In our previous study [17], we observed that the classification performances of MI EEG signals were evaluated on the BCI Competition III dataset [18] and IV dataset [19] by using traditional neural network [20-30] and DNNs [17, 31-47]. Actually, classification performance for four-class MI EEG signals is still not at a high level. Fortunately, DNNs produce high classification performances in almost all machine learning areas. Besides their benefits, the DNNs have the following two major drawbacks among the others in classification of MI EEG signals: (i) DNNs need big datasets, and (ii) determination of even a coarse structure for the DNN may take days. We have noticed that by solving some problems encountered in DNNs, classification performance for the MI EEG signals can be improved. A solution for increasing the classification performance on small-sized MI EEG datasets was to use transfer learning [41]. In the transfer learning of [41], the huge network pre-trained using a different and probably big data set is retrained by the BCI MI EEG dataset. Though, high successes were achieved, the DNN really had excessive number of nodes. In another solution, researchers mostly preferred using preprocessing or transformation stage preceding the CNN structure to provide high classification performances with small

datasets [31-43,44-46]. The CSP [31,32,34,35,40,43,44-47], FFT [33], STFT [38,39] or CWT [41] were preferred for the preprocessing (or transformation) stage. In those studies, CNNs investigate the features in the input space determined by the preprocessing. However, these stages introduce considerable amount of computational load to the decision making system. Because the classification performance is very dependent on the correct determination of the parameters of the preprocessing stage, in addition to a successful training of the DNN, the researchers need to determine the optimum parameters for each subject.

Determining pneumonia from chest x-ray (CXR) images is an extremely difficult and important image processing problem. The discrimination of whether pneumonia is of bacterium or virus origin has also become more important during the pandemic. Automatic determination of the presence and origin of pneumonia is crucial for speeding up the treatment process and increasing the patient's survival rate. Deep neural networks (DNN), popular in recent years, are widely used in CXR image classification to detect pneumonia, and high performance is provided [48-64]. There are several databases of CXR images. The studies [49-55] using other databases, and studies [48, 49, 56-64] using the same database [65] with this study have classified CXR images into two broad categories: normal and pneumonia. Only Kermany et al. [64] has classified as bacterial or viral pneumonia, as in our study [48], in addition to the classification of normal or pneumonia. It has been observed that the augmentation process for CXR images in [51, 55-57, 59, 62, 64] and preprocessing stages in [48, 49, 51-53, 55, 56, 59-64] were not preferred. Transfer learning has been used to strengthen the training of DNNs, especially when the dataset is of small size. In fact, the size of datasets used for CXR images is not small. It is observed that DNNs using transfer learning contain excessively large number of weights and also do not give higher classification performance than DNNs trained with random weights. Moreover, in some studies [52, 56, 57, 60, 63], CNN plus FCNNs are preferred to classify the CXR images. This also leads DNN structure which have the excessively large number of weights.

We also observed [66] that researchers have proposed series of image processing and pattern recognition algorithms for the computer-aided inspection and classification of solder joints in BGA X-ray images. In the literature, studies inspecting the conditions of solder joints generally categorize the BGA X-ray images into two, three or four broad classes [67-69]. In the majority of these studies, morphological image processing methods are used for the analyses, and results are given for a few images. And, in some of these studies classification performances are not mentioned. In our previous study [66], the proposed DNN model contained feature extractor layers and a minimum distance classifier. Since the proposed network in [66] consisted of less number of layers (four convolution layers and one fully connected layer), determination of the hyper-parameters of the network and training of the network were accomplished in a short time. BGA X-ray images were categorized into four classes according to the conditions of the solder joints: normal, short-circuit, bonding defect and void defect.

In our previous studies, ECG [8], EEG [17], heart sound, the pneumonia in X-ray chest images [48], BGA solder defects [66], and patterns of benchmark datasets (MNIST, IRIS, CIFAR10 and CIFAR20) [8] were classified with high success. In these studies, the goal was to strengthen the DNN's training and thus achieve high classification performances for above mentioned patterns using a small-sized network. The DNN structure proposed in the studies [8, 17, 48, 66] was gradually developed, and it reached its final form [17]. In this study, applications for developed versions of DivFE and the reached final form will be discussed in detail.

In general, DNNs consist of two parts: a CNN (feature extractor), and a fully connected neural network (FCNN) or another neural network such as the LVQ (linear vector quantization) and SVM, among others. In some cases, the DNN contains either the CNN or the FCNN. In general, the FCNN owns the majority of the nodes of DNN. In this study, the proposed DNN is comprised of a CNN and a minimum distance network as the classifier instead of the FCNN. Therefore, the proposed method focuses solely on the training of the CNN, as the minimum distance network does not require any training. Training both the convolutional layer and classifier (such as fully connected layers) of deep networks simultaneously for each input and output pair makes it difficult to determine the weights of the network optimally. Because, the risk of being caught by the local optima increases as the number of nodes of the network takes high values. In these studies [8, 17, 48, 66], with the developed training strategy, it was possible to train the convolutional layers (feature extractor) and classifiers individually. If both are desired to be included in the network, first the convolutional layer and then classifier (such as the fully connected layer) can be trained sequentially. If they are trained separately, the features are more likely to be accurately determined. As a result of the achievements in the extracted features, the fully connected neural layers can be eliminated. Therefore, we preferred to employ only a simple minimum distance classifier instead of a fully connected layer.

## 2. METHODOLOGY

### 2.1 Structure of the proposed convolutional neural networks

In general, a traditional DNN structure comprises of two cascaded units: a convolutional neural network which in effect plays the role of the *feature extractor* (FE), and a fully connected neural network as the *classifier*. The CNN structure is a combination of the convolutional layer, pooling layer, and rectified linear units (ReLU) layer.

The convolutional layer is the major processing layer in the CNN architecture which has a set of kernels (learnable filters) with small receptive fields. These filters are convolved with the input signal resulting in a 1D activation map for each filter. The pooling layer in the CNN performs a nonlinear down-sampling operation, and in the meantime, contributes to the translation invariance of the network. By inserting a pooling layer between the CNN's convolutional layers, the input size of the CNN is progressively reduced. Among several nonlinear functions, max-pooling is the most preferred pooling operation in machine learning applications. The need for using the pooling layers varies according to the problem and there is no general methodology. In the literature, pooling process is widely used in CNNs, but, this layer does not need to be used after every convolutional layer. In this study [17], a pooling layer is used between all the successive convolutional layers. With an activation function of $f(x)=\max(0, x)$, the ReLU layer sets the negative values in the activation maps to zero, thus performs a nonlinear operation in the decision making system. Moreover, the batch normalization and drop-out functions are used between each layer to further strengthen the training. While the CNN generally represents the features, the FCNN represents the classifier. It is difficult to determine the hyper-parameters of CNN and FCNN, and to train both structures simultaneously. When these two parts of DNN are employed, the number of hyper-parameters of DNN increases drastically; some of these parameters are (*i*) number of convolutional layers, (*ii*) number of feature planes in convolutional layers, (*iii*) filter size, (*iv*) number of hidden layers in FCNN, and (*v*) number of nodes in the hidden layers. It may take days to determine the optimum values of these parameters. If the DNN consists of only the CNN, the number of hyper-parameters of DNN decreases to (*i*) number of convolutional layers, (*ii*) number of feature planes in convolutional layers, and (*iii*) filter size, *etc*.

In our study [17], the proposed DNN is comprised of the CNN, and a minimum distance network (MDN) -instead of the FCNN- as the classifier. The proposed network provided in Fig. 1 is called DivFE which is the acronym for the *Divergence-based Feature Extractor*.

The FCNN and its training phase introduces a lot of problems, some of which are (*i*) failing to converge to the optimum solution, (*ii*) high computing load in real time applications, (*iii*) high memory requirements in portable systems (such as FPGA or embedded Linux cards), and (*iv*) determination of the hyper-parameters such as the number of hidden layers and the number of nodes in these hidden layers.

After several convolutional layers, a MDN is used to determine the class (label) of the input signal. The weights of the nodes of the MDN do not change during the training. These weights are set to the values in the rows (or columns) of the Walsh matrix. The output of the proposed feature extractor is fed to the input of the MDN. The node of the MDN that is the closest to the FE's output determines the class of the input signal. MDN determines the label of the class of the input signal by using the equations

below:

$$D_k = \sum_{j=1}^{M}(O_j - H_{k,j})^2 \qquad D_i = \min_k(D_k) \qquad (1)$$

where $M$ represents the dimension of the Walsh matrix (in this study $M$ is equal to 16), $O_j$ represents the $j$th output of the feature extractor (and also outputs of the flatten layer in Fig. 1), $H_{k,j}$ represents the $j$th element of the Walsh vector belonging to the $k$th class, index $i$ represents the label of the class which is the decision of the MDN.

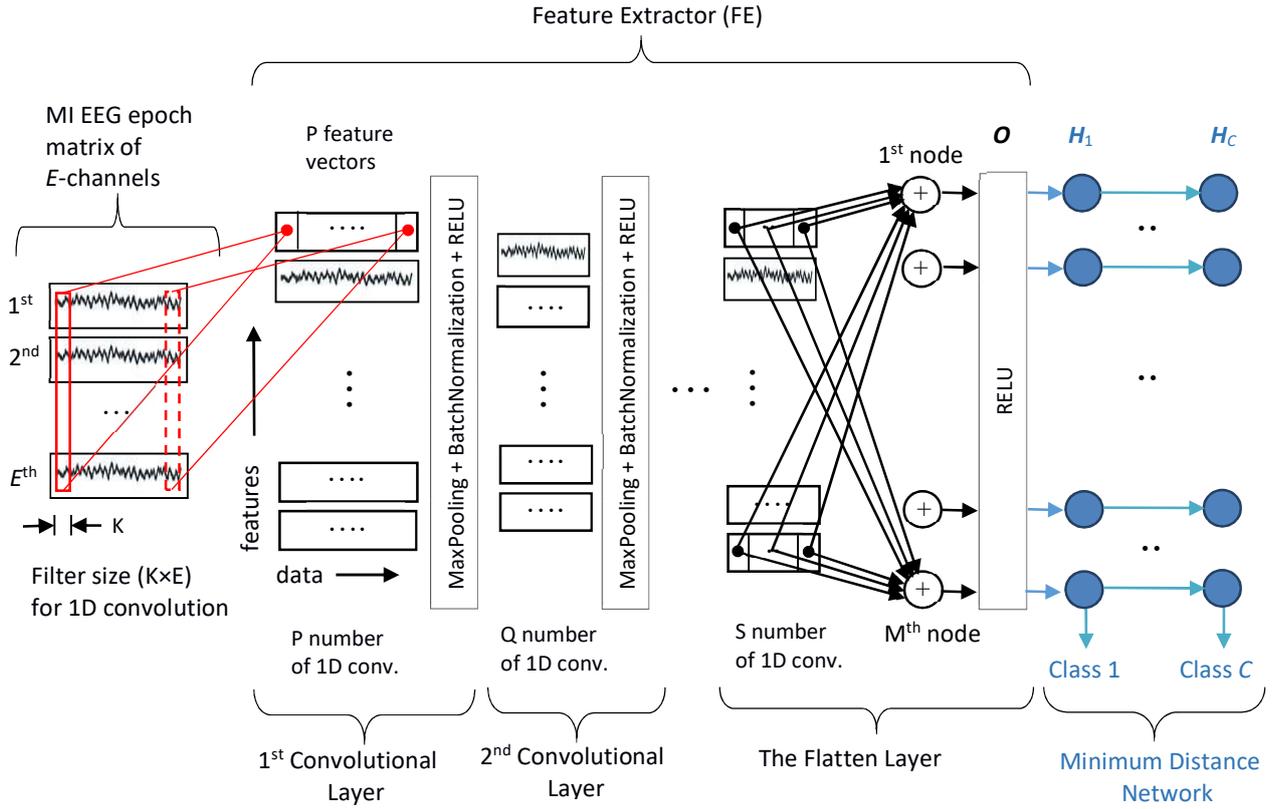

**Fig. 1** The structure of the DivFE [17]. $O$ represents the flatten layer. The size of the $O$ vector is equal to the size of Walsh vectors. The structure of the Feature Extractor is not different from the CNN known in the literature. In this architecture, a minimum distance network (MDN) is proposed instead of the fully connected neural network. $H_k$ is $k$th node of the MDN constituted with the rows of Walsh matrix. The values of $H_k$ are not updated in the training.

## 2.2 Training the convolutional neural networks

Training both the convolutional layer and the classifier (such as fully connected layers) of deep networks simultaneously for each input and output pair makes it difficult to determine the weights of the network optimally. Because, the risk of being caught by the local optima increases as the number of nodes of the network takes high values. In this study [17], with the developed training strategy, it is possible to train the convolutional layer and the classifier (such as fully connected layers) individually. If both are desired to be included in the network, first the convolutional layer and then the classifier (such as fully connected layer) can be trained sequentially. If they are trained separately, the features are more likely to be accurately determined. As a result of the achievements in the extracted features, the fully connected neural layers can be eliminated. Therefore, we preferred to employ only a simple minimum distance classifier instead of a fully connected layer.

In the developed training strategy, it is aimed to achieve a high divergence value for the feature space searched by the CNN layer. In classification problems, the divergence value computed over a feature set, points to the effectiveness of the features in terms of being intra-class representative and inter-class discriminative. In this respect, a class separability definition is made as given in the following equation:

$$\text{divergence value} = \text{tr}(\boldsymbol{S}^{-1} \cdot \boldsymbol{B})$$
$$\boldsymbol{S} = \boldsymbol{S}_1 + \cdots + \boldsymbol{S}_k + \cdots + \boldsymbol{S}_C \qquad (2)$$

Here, $\text{tr}(\cdot)$ is the trace operation, $\boldsymbol{S}_k$ is the covariance matrix of the $k$th class, and the sum of the covariance matrices of all the $C$ classes is denoted by $\boldsymbol{S}$ which is also called the within-class scatter matrix of the distribution. The $\boldsymbol{B}$ is the between-class scatter matrix, and it is the covariance of the mean vectors of the classes. $\boldsymbol{S}$ matrix is calculated by using output vectors $\boldsymbol{O}$ of FE. According to Eq. (2), the ratio of a larger between-class scatters to a smaller within-class scatters leads to a high divergence value to be obtained. A high divergence value signifies that a favorable distribution of the feature vectors has been attained.

In this regard, to distribute the mean vectors (centers) of the classes at the farthest positions in the feature space with equal distances from each other, the Walsh vectors showed up to meet these requirements. By associating each row (or column) of the Walsh matrix with a class center, given an input vector, the FE is trained to output the specific row (or column) of the Walsh matrix which represents the class label of that input vector. With this training strategy, selecting the output vectors of the FE as the rows of the Walsh matrix increases the distances between the class centers, thus increases the divergence value which is an indication of efficient features. The $\boldsymbol{B}$ matrix is calculated by using the $\boldsymbol{H}_k$ vectors.

To realize the ReLU function in the FE, 0 and 1 values had to be used in place of −V and V in the Walsh matrices, respectively. In Eq.(3), the original Walsh matrices for two- and four-dimensional spaces and a modified Walsh matrix for an eight-dimensional space are given.

$$H = \begin{bmatrix} +V & +V \\ +V & -V \end{bmatrix} \quad H = \begin{bmatrix} +V & +V & +V & +V \\ +V & -V & +V & -V \\ +V & +V & -V & -V \\ +V & -V & -V & +V \end{bmatrix} \quad H = \begin{bmatrix} 1 & 1 & 1 & 1 & 1 & 1 & 1 & 1 \\ 1 & 0 & 1 & 0 & 1 & 0 & 1 & 0 \\ 1 & 1 & 0 & 0 & 1 & 1 & 0 & 0 \\ 1 & 0 & 0 & 1 & 1 & 0 & 0 & 1 \\ 1 & 1 & 1 & 1 & 0 & 0 & 0 & 0 \\ 1 & 0 & 1 & 0 & 0 & 1 & 0 & 1 \\ 1 & 1 & 0 & 0 & 0 & 0 & 1 & 1 \\ 1 & 0 & 0 & 1 & 0 & 1 & 1 & 0 \end{bmatrix} \quad (3)$$

In the modified Walsh matrix, it can easily be noticed that the Hamming distance between any two rows or columns is equal to the half of the matrix rank value (rank is equal to the dimension of output vectors). Therefore, the nodes of the MDN are chosen from the rows (or columns) $H_k$ of the modified Walsh matrix. In this study [17], Walsh matrix rank was chosen as 16 for both the two- and four-class problems. The first two and four rows in the Walsh matrix are reserved to represent the class centers (nodes) of MDN for two- and four-class problems, respectively. Increasing the matrix rank also increases the Hamming distances between the class centers, contributing to the class separability criterion. On the other hand, over increasing the rank should be avoided as it causes prolonged training phases. Fig. 2 shows the training algorithm of the DivFE.

Walsh functions are preferred due to the following four properties: (*i*) It is known that in any feature space to satisfy the divergence criterion at its best, the class centers should locate distantly from each other as much as possible, and within-class scatterings should be small. Since the nodes of the MDN hold the centers of the classes, by representing each class center by a row (or column) of the Walsh matrix, the centers of the classes distribute over the feature space at the farthest possible distances. This approach increases the divergence value. (*ii*) In general, increasing the number of features (the number of weights in the nodes of MDN) improves the classification performance. In the proposed method, the dimension of the feature space can be set to high values by simply increasing the rank of Walsh matrix. This approach will provide a better class distribution in a multi-dimensional space. (*iii*) In deep learning, the best performing features and the classifier weights are both searched simultaneously during the training phase. Since, convergence problems may happen with such dense networks, in the proposed method, it is suggested to train the CNN (feature extractor) and the classifier individually. In this study [17], MDN is preferred as a classifier. By copying the content of the Walsh matrix to the nodes of the MDN, the classifier is made ready to be used without an iterative training algorithm. Hence, during the training, only the CNN is trained to map the input vectors to Walsh vectors, and the MDN is only run during the testing phase. Because the FCNN was not used, the training algorithm was focused only on the features, so eventually, a decrease in the within-class scatterings was achieved. (*iv*) Transfer learning (the use of pre-trained networks) has also become a popular choice for the applications with deep learning. A pre-trained network designed with a Walsh matrix of rank greater than the number of classes can easily be expanded to learn the distribution of the new additional classes. In this studies, as the rank of Walsh matrix was 16 for both the two- and four-class problems studied, if requested, the pre-trained network can be retrained for the additional 14 or 12 new classes, respectively. In the proposed study the dimension of the feature space is not kept too large in order to obtain the classification results quickly, and to make the network learn the dataset quickly.

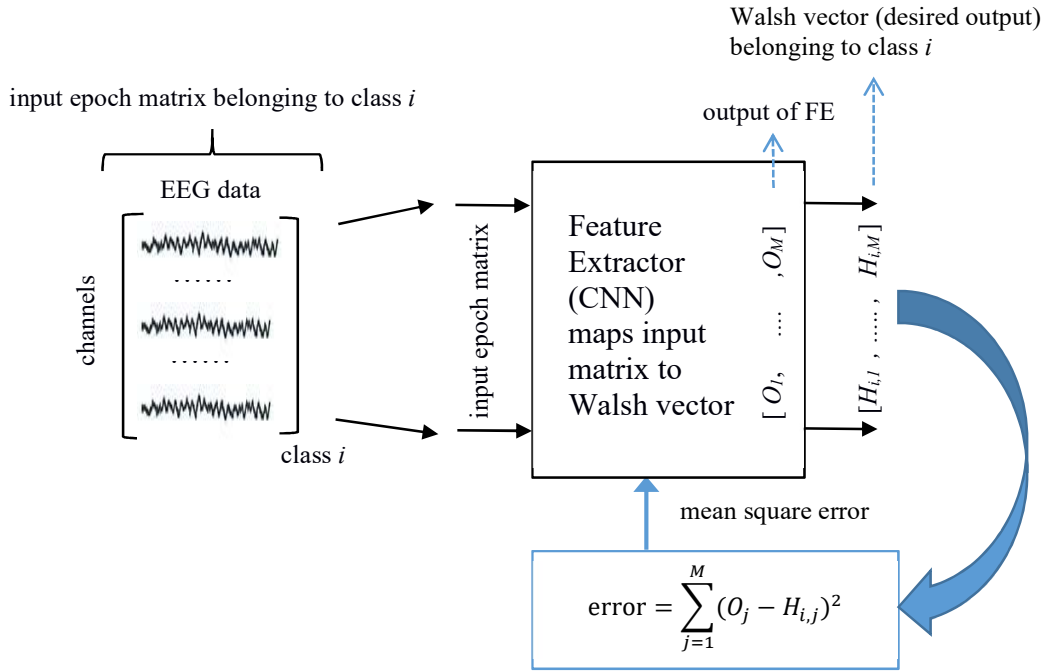

Fig. 2 Training process of DivFE [17]. $O_j$ represents the *j*th output of the feature extractor (and also the outputs of the flatten layer in Fig.1), and $H_{i,j}$ represents the *j*th element of Walsh vector belonging to the *i*th class. In this study *M* is equal to 16. As an example, for two classes:
$\boldsymbol{H}_1$ =[1,0,1,0,1,0,1,0, 1,0,1,0,1,0,1,0] and $\boldsymbol{H}_2$ =[1,1,0,0,1,1,0,0,1,1,0,0,1,1,0,0]
(Except $\boldsymbol{H}_i$ = [1,1,1,1,1,1,1,1,1,1,1,1,1,1,1,1]).

**2.3 Dataset preparation and validation process**

The differences between the training and validation processes, and the structures of a basic DNN and DivFE are pointed out in Algorithm 2. Generally, 10% of the training set is selected as the validation set. After all the training data is used for once for the training of FE (it is known as one iteration), accuracy and mean squared error loss values are calculated for all data in the validation set. In the validation phase, each input data (epoch) of the validation set is given to the FE (shown in Fig. 1), and FE generates an output vector for each input data. This output vector $\boldsymbol{O}$ is presented to the MDN. The node of the MDN that is the closest to the $\boldsymbol{O}$ determines the label of the class for the input data as the decision. After the assigned Walsh vectors (class labels) for each input data are compared with the vectors generated by the MDN, mean accuracy and loss values are calculated over all the data in the validation set.

**Algorithm 2:** Differences between the training and validation processes, and the structures of a basic DNN and DivFE for two classes [17]

**Basic DNN:** Feature Extractor (FE convolutional neural network) + Classifier (FCNN)
**1- One iteration of training process for two classes**

**for** each input data in {training set} **do**
   **run** the network (FE + FCNN) and obtain the *network output* $\boldsymbol{O} = [O_1, O_2]$
   **define** *desired output*: $[0, 1]^T$ for the input data of the first class
                               $[1, 0]^T$ for the input data of the second class
   **update** weights of network (FE+ FCNN) by calculating $\sum(network\ output - desired\ output)^2$
**end for**

**2- Validation/Test process for two classes**

**for** each input data in {validation/test set} **do**
   **run** the network (FE + FCNN) and obtain the *network output* $\boldsymbol{O} = [O_1, O_2]$
   **obtain** the decision "$i$" of basic DNN by calculating the below equation

$$O_i = \max(O_1, O_2)$$
**end for**
**calculate** accuracy by comparing the decision of basic DNN for each input data with the class label of the input

**DivFE** shown in Fig. 1: Feature Extractor (FE convolutional neural network) + Classifier (MDN)
**1- One iteration of training process for two classes**

**for** each input data in {training set} **do**
   **run** the network (FE) and obtain the *network output* $\boldsymbol{O}$
   **define** *desired output*: $\boldsymbol{H}_1 = [1,0,1,0,1,0,1,0,1,0,1,0,1,0]^T$ for the input data of the first class
                            $\boldsymbol{H}_2 = [1,1,0,0,1,1,0,0,1,1,0,0,1,1,0,0]^T$ for the input data of the second class
                            Except $\boldsymbol{H}_i = [1,1,1,1,1,1,1,1,1,1,1,1,1,1,1,1]^T$
   **update** weights of network (FE) by using $\sum(network\ output - desired\ output)^2$
**end for**

**2- Validation/Test process for two classes**

**for** each input data in {validation/test set} **do**
   **run** the network (FE) and obtain the *network output* $\boldsymbol{O} = [O_1, \ldots, O_{16}]$
   **obtain** the decision "$i$" of DivFE by calculating the below equation for $\boldsymbol{H}_1$ and $\boldsymbol{H}_2$

$$\text{MDN:} \quad \boldsymbol{D}_k = \sum_{j=1}^{M}(O_j - H_{k,j})^2 \quad \boldsymbol{D}_i = \min_k(\boldsymbol{D}_k)$$
**end for**
**calculate** accuracy by comparing decision of DivFE for each input data with the class label of the input

While determining a coarse structure for the FE, mean squared error loss values and accuracies for both the training and validation sets are examined to avoid under-fitting and over-fitting. During this stage, hyper-parameters (filter size, number of layers, number of features in the layers, *etc*.) of the FE are determined by trial and error, as done in the literature. After the coarse structure is determined, FE's training is started for long iterations. The training is terminated considering both the iteration number and the loss value for the validation set. And then, the accuracy for the test set is calculated. For *N* randomly generated training and test sets, the FE is trained *N* times and the average of *N* accuracies for the test sets is calculated to achieve statistically significant success.

## 3. COMPUTER SIMULATIONS

All the analyses were realized using Python codes running on Ubuntu Linux workstation which had 32 core CPUs of 2.7 GHz with GeForce GTX1080 Ti Graphics card.

In this study, DivFE's capability was investigated by applying it to problems in different areas. In this context, performances of the DivFE in classification of ECG, EEG, heart sounds, pneumonia in X-ray chest images, BGA solder defects, and patterns of benchmark datasets (MNIST, IRIS, CIFAR10 and CIFAR20) are discussed.

### 3.1 Comparative results in the classification of ECG signals and ECG images

In our previous study [8], aim was to develop an ECG classification method which provides high accuracy and fast results without using extra features and complex operations in portable systems (especially, using the camera of a mobile system). In this context, major design criteria were: to classify a large number of different types of ECG signals, to achieve high performance in the classification process, not to use extra parameters (such as R-R intervals or Fourier transform) in the classification process, and to reduce the number of nodes of the CNN network in order to produce fast results.

The MIT-BIH arrhythmia database has been used as the standard ECG database in almost all of the studies [1-16] in the classification of ECG signals using convolution neural networks. This database was employed in this study as well, because it contains a large number of beat types: Normal beat (N), left bundle branch block beat (L), right bundle branch block beat (R), atrial premature beat (A), aberrated atrial premature beat (a), premature ventricular contraction (V), fusion of ventricular and normal beat (F), ventricular flutter wave (O), ventricular escape beat (E), paced beat (P), non-conducted p wave (blocked APB - p), etc. In this subsection, comparisons have been made in three categories: (*i*) classification of 1D ECG signals using CNNs, (*ii*) classification of 1D ECG signals using other classifiers, and (*iii*) classification of ECG images using CNNs.

Tables 1-2 summarize the methods in terms of the database used in the studies, the number of classified beat types (number of classes), extra features or processing stages which are needed for the classification, average test classification performance, and the number of weights used in the networks.

### 3.1.1 Comparative results for the classification of 1D ECG signals

In Table 1, it is observed that the numbers of weights of the networks shown in the first three rows [1-3] are low compared with the other studies, because these networks are employed to classify only two different ECG beat classes. As can be seen from Table 1, it is natural that the number of nodes in the network increases as the number of classes increases. The classification performance has been improved by using extra features carrying more information [1-3] in addition to the features directly formed with the ECG amplitudes. Since the network developed in our study [8] does not have a fully connected layer, the main focus in the training is on searching for the best features by the aid of convolution layers. Hence as a result, the features were found to be more accurate and for this reason 11 different ECG beat types are classified with high performance using a small number of nodes. It is known that it is difficult to achieve high classification performances as the number of classes increases. This can be observed in [4, 5] that when the number of classes was increased to five, although both of the convolution and fully connected layers have been used together, high success rates could not be achieved.

**Table 1** Comparison results of the classification of 1D ECG signals by using deep learning [8]

| Methods | Database | Number of classes | Extra features or processing stages | Number of weights (#CoL , #FCL) #WCoL + #WFCL | Average test classification accuracy (%) |
|---|---|---|---|---|---|
| Sannino et al. [1] | MIT-BIH | 2 | R-R interval values | (0 , 6) 0 + 3970 | 99.09 |
| Rahhal et al. [2] | MIT-BIH | 2 | R-R interval values | (0 , 2) 0 + 11600 | 98 |
| Jun et al. [3] | MIT-BIH | 2 | Six different features | (0 , 7) 0 + 11700 | 99 |
| Acharya et al. [4] | MIT-BIH | 5 | No | (3 , 3) 1015 + 18700 | 93.4 |
| Kachuee et al. [5] | MIT-BIH | 5 | No | (11 , 2) 51360 + 5280 | 95.9 |
| In our study [8] | MIT-BIH | 11 | No | (4 , 0) 26820 + 0 | 99.45 |

CoL : Convolution layers     WCoL : Weights in convolution layers
FCL : Fully connected layers     WFCL : Weights in fully connected layers

In Table 2, the performances of the classifiers other than the deep networks for 1D ECG signals are presented. It is observed in these studies that high classification performances were obtained by using complex processes. However, complex processes will cause the memory requirements and computational load to increase and hence, will limit the practical implementation of these methods on portable mobile systems. As can be observed in Table 1, similar performance rates were achieved using deep learning by adding extra features [1-3] or not including extra features as in our study.

**Table 2** Comparison results of the classification of 1D ECG signals by using other classifiers [8]

| Methods | Database | Number of classes | Complexity: feature extraction or preprocessing | Average test classification accuracy (%) |
|---|---|---|---|---|
| Jiang et al. [9] | MIT-BIH | 2 | Hermite basis functions | 98.15 |
| Ince et al. [10] | MIT-BIH | 2 | WBMF + PSO | 96.85 |
| Kallas et al. [11] | MIT-BIH | 3 | KPCA + SVM | 97.39 |
| Balouchestani et al. [12] | MIT-BIH | 5 | PCA + KNN | 99.9 |
| Chazal et al. [13] | MIT-BIH | 5 | RR interval + Morphology features + statistical classifier | 86.2 |
| Zhang et al. [14] | MIT-BIH | 4 | Preprocessing + RR measures + Morphology features + SVM | 88.34 |
| Martis et al. [15] | MIT-BIH | 3 | DCT +PCA DWT+PCA | 99.45 |
| Li et al. [16] | MIT-BIH | 5 | Random Forest + DWT + RR measures | 94.61 |
| In our study [8] | MIT-BIH | 11 | No extra features or processing stages | 99.45 |

WBMF : Wavelet-based morphological features  
PSO : Particle swarm optimization  
KPCA : Kernel principal component analysis  
SVM : Support vector machines  
LCC : Linear correlation coefficient  
PNN : Probabilistic neural network  
K-NN : K-nearest neighbors  
DCT : Discrete cosine transform  
DWT : Discrete wavelet transform  

### 3.1.2 Conversion from 1D biological signals to images in the classification process

There is an emerging research on applying deep learning techniques to the converted signals from one dimension to two dimensions. In this context, it is observed that there are many studies in the literature for the classification of heart sounds, respiratory sounds, electroencephalography signals (EEG), ECG signals and limb movements [6-7]. In these studies, the researchers reported that they had achieved high classification performances. Some of the studies that gave high performances in the classification of ECG images were shown in Table 3.

As observed in Table 3, high performances were achieved in classifying ECG signals without using complex procedures. Converting the 1D ECG signals to images increased the classification performance by about 5 to 6 points compared with the studies in Table 1 [4-5]. In the studies in [4, 5], no extra features or processing stages were used. The researchers in [6, 7] took advantage of having specific ECG waveforms in two dimensions. Moreover, it is observed that the increase in the number of nodes of networks usually comes from the fully connected layers. In the flatten layer of the network proposed in Wu's study [6] there are $28 \times 28 \times 128$ nodes. This value is multiplied by the number of nodes in the first fully connected layer (2048) to compute the number of weights ($28 \times 28 \times 128 \times 2048$) in this layer. There are $2048 \times 2048$ weights in the second fully connected layer. And finally, there are $2048 \times 8$ weights in the third fully connected layer.

**Table 3** Comparison results of the classification of the ECG images by using deep learning [8]

| Methods | Database | Number of classes | Extra features or processing stages | Number of weights (#CoL, #FCL) #WCoL + #WFCL | Average test classification accuracy (%) |
|---|---|---|---|---|---|
| Wu et al. [6] | MIT-BIH | 2 | No | (5, 3) 421408 + 209719296 | 98 |
| Jun et al. [7] | MIT-BIH | 8 | No | (9, 1) 1143360 + 134234112 | 98.9 |
| AlexNet used in Jun et al. [7] | MIT-BIH | 8 | No | (6, 3) 833403 + 256131072 | 98.8 |
| VGGNet used in Jun et al. [7] | MIT-BIH | 8 | No | (10, 3) 7631424 + 71319552 | 98.7 |
| In our study [8] | MIT-BIH | 11 | No | (5, 0) 212892 + 0 | 98.7 |

CoL : Convolution layer     WCoL : Weights in convolution layers
FCL : Fully connected layer     WFCL : Weights in fully connected layers

### 3.2 The classification of MI EEG signals

In our previous study [17], the best classification performances of MI EEG signals were obtained on the BCI Competition III datasets IIIa and IVa [18], and BCI Competition IV datasets IIa and IIb [19] by using the DivFE. BCI Competition III datasets IIIa and IVa have MI EEG records from three and five subjects, respectively. BCI Competition IV datasets IIa and IIb each has MI EEG records from nine subjects. Two different models were compared: (i) only the proposed DNN (without the CSP), and (ii) the CSP + proposed DNN. We have investigated the effects of augmentation process on the classification performance of MI EEG for each model. We have evaluated these models on 4 different datasets (BCI Competition III-IIIa, III-IVa, IV-IIa and IV-IIb).

Four different experiments were covered in order to examine the effects of the augmentation and transformation on the classification of MI EEG signals: (*i*) no augmentation + no CSP, (*ii*) no augmentation + CSP, (*iii*) augmentation + no CSP, and (*iv*) augmentation + CSP. These experiments were applied to four different datasets. First of all, 80% of each dataset was allocated for the training and the remaining 20% for the test. If the augmentation process will be used in the experiments, a new training set is formed by augmenting the epochs in the existing training set.

Table 4 shows the classification results obtained with traditional neural networks by using the benchmark datasets [18, 19]. As can be observed from the table, high success rates could not be achieved with traditional neural network topologies, and the average performances decreased when the number of classes was increased. Since the classification performance is highly dependent on the determination of efficient features, researchers have worked intensively on the feature extraction methods to determine the optimum features for each subject.

**Table 4**: The classification of MI EEG signals with the traditional neural networks. [17]

| methods | accuracy % | number of classes | transformation + feature extraction | classifier | dataset |
|---|---|---|---|---|---|
| Wang's method [20] | 77.2 | 4 | CSP + variance | MLP | BCI III-IIIa |
| Aljalal's method [21] | 80.2 | 2 | Wavelet + statistical, entropy, energy features | MLP | BCI III-IVa |
| Mirnaziri's method [22] | 61.7 | 4 | CSP + variance | MLP | BCI IV-IIa |
| Silva's method [23] | 67.8 | 2 | Linear Predictive Coding | MLP | BCI IV-IIb |
| Alansari's method [24] | 83.8 | 2 | Wavelet | SVM | BCI IV-IIb |
| Behri's method [25] | 89.4 94.5 67.4 | 2 | Wavelet | SVM K-NN MLP | BCI III-IVa |
| Zhang's method [26] | 84 | 2 | CSP + variance | SVM | BCI III-IVa |
| Li's method [27] | 68.6 | 3 | FBCSP+ variance | SVM | BCI IV-IIa |
| Wang's method [28] | 81.2 | 2 | FD-CSP + variance | SVM | BCI IV-IIb |
| Mishuhina's method [29] | 89.8 | 4 | RCSP – FWR | LDA | BCI III-IIIa |
| Molla's method [30] | 92.2 91.3 91.2 | 2 | CSP + subband features | SVM LDA K-NN | BCI III-IVa |

Recent studies on the classification of MI EEG signals by using the DNNs were summarized in Table 5. The four rows at the bottom of the table show the classification results obtained by the proposed methods in our study [17].

In our study [17], the advantages of the augmentation process were also demonstrated. For the cases when transformation was applied to the raw data, there was a slight difference between the classification performances obtained with augmented and non-augmented training data. But it was observed that if transformation is not used, there is a huge difference between the classification results produced by the augmentation and non-augmentation processes. Two sets were created with the classification results obtained from 26 subjects for augmentation and non-augmentation processes in the case of applying no transformation. The paired t-test was calculated to demonstrate the effect of using the augmentation process. The violin plot provided in Fig. 3 allows for a visual investigation of the effect of augmentation process on the accuracy of the classification results. It is clear that, in case of augmentation the distribution of the achieved accuracy results is skewed with a mode value of 98.2%, whereas in case of non-augmentation the distribution is almost symmetric with a mode value of 67.2%. A paired t-test further validated that, the obtained accuracy results with augmentation process is statistically significantly higher (p-val: 2.33e−12) than the results obtained with non-augmentation process.

**Table 5** Classification results of DNNs for MI EEG signals [17].

| studies | database-dataset | number of classes | accuracy % | transformation | number of subjects |
|---|---|---|---|---|---|
| Yang's study[31] | IV-IIa | 4 | 69.27 | Filter + CSP | 9 |
| Sakhavi's study [32] | IV-IIa | 4 | 70.6 | Filter + CSP | 9 |
| Lu's study [33] | IV-IIb | 2 | 84 | FFT | 9 |
| Sakhavi's study [34] | IV-IIa | 4 | 74.46 | Filter + CSP | 9 |
| Abbas's study[35] | IV-IIa | 4 | 70.7 | Filter + CSP | 9 |
| Car's study[36] | IV-IIa | 4 | 70.5 | no transformation | 9 |
| Wu's study[37] | IV-IIb | 2 | 80.6 | Filter bank | 9 |
| Dai's study[38] | IV-IIb | 2 | 78.2 | STFT | 9 |
| Tabar's study[39] | IV-IIb | 2 | 77.6 | STFT | 9 |
| Thang's study[40] | III-IIIa | 4 | 91.9 | CSP | 3 |
| Chaudhary's study[41] | III-IVa | 2 | **99.3** | CWT | 5 |
| Zhao's study[42] | IV-IIa | 4 | 75 | 3D-EEG | 9 |
| Zhang's study [43] | IV-IIb | 2 | 82 | CSP+bispectrum | 9 |
| Deng's study [44] | III-IIIa | 4 | 85.3 | FBCSP | 3 |
|  | IV-IIa | 4 | 78.9 | FBCSP | 9 |
| Olivas-Padilla's study[45] | IV-IIa | 4 | 78.4 for monolithic network | DFBCSP | 9 |
| Liu's study [46] | IV-IIa | 4 | 76.86 | CSP | 9 |
| Soman's study [47] | III-IIb | 2 | 76.3 | CSP | 9 |
| NTS-A-III/IVa | III-IVa | 2 | 96.2 | no transformation | 5 |
| TS-A-III/IVa |  |  | 98.5 | CSP |  |
| NTS-A-III/IIIa | III-IIIa | 4 | **96.5** | no transformation | 3 |
| TS-A-III/IIIa |  |  | 95.8 | CSP |  |
| NTS-A-IV/IIa | IV-IIa | 4 | **79.3** | no transformation | 9 |
| TS-A-IV/IIa |  |  | 79.1 | CSP |  |
| NTS-A-IV/IIb | IV-IIb | 2 | **88.6** | no transformation | 9 |
| TS-A-IV/IIb |  |  | 85.1 | CSP |  |

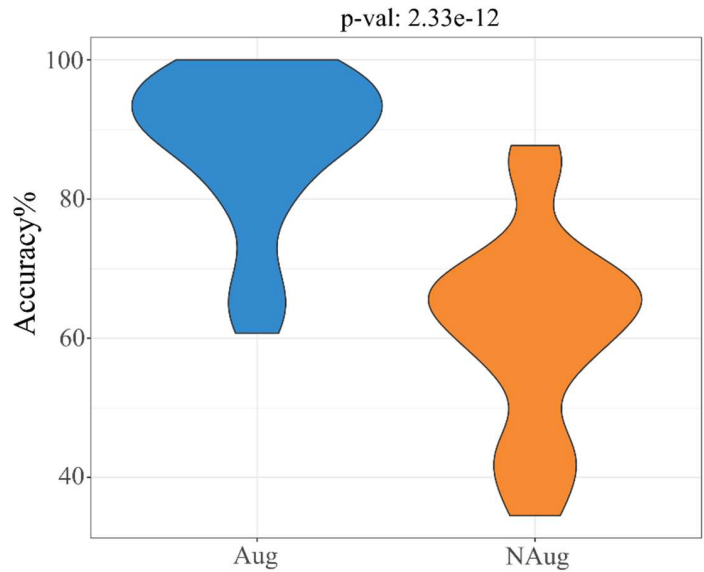

Fig 3. The paired t-test to demonstrate the effect of using the augmentation process [17].

**3.3 Detection Pneumonia in X-ray Chest Images**

Deep neural networks (DNN), popular in recent years, are widely used in CXR image classification to detect pneumonia, and high performance is provided [48-64]. CXR images were selected among pneumonia patients at the Guangzhou Women and Children's Medical Centre (GWCMC) and CXR images were classified into two categories: normal and pneumonia. The used dataset [65] in which images of Chest X-ray (shown in Fig. 4) have been labeled as normal and pneumonia, consists of Chest X-ray images from one to five year old children provided by the Guangzhou Women's and Children's Health Center. Of these images, low-quality or illegible ones have been eliminated and the remaining images have been graded by a total of three experts for use in artificial intelligence studies. There are 5840 images in total, each of which is different in size. A total of 5,232 chest X-ray images is collected from children, including 3,883 characterized as depicting pneumonia (2,538 bacterial and 1,345 viral) and 1,349 normal, from a total of 5,856 patients to train the AI system. The model was then tested with 234 normal images and 390 pneumonia images (242 bacterial and 148 viral) from 624 patients.

In our previous study [48], three different DNNs were developed to classify the chest X-Ray images: (1) as normal or pneumonia (DNN_NP), (2) as pneumonia viral or pneumonia bacterial (DNN_VB) and (3) as normal, pneumonia viral or pneumonia bacterial (DNN_NVB). In all experiments, Chest X-ray images were entered to the DNNs without be enriched the training set by using augmentation and without a preprocessing stage applied to input images.

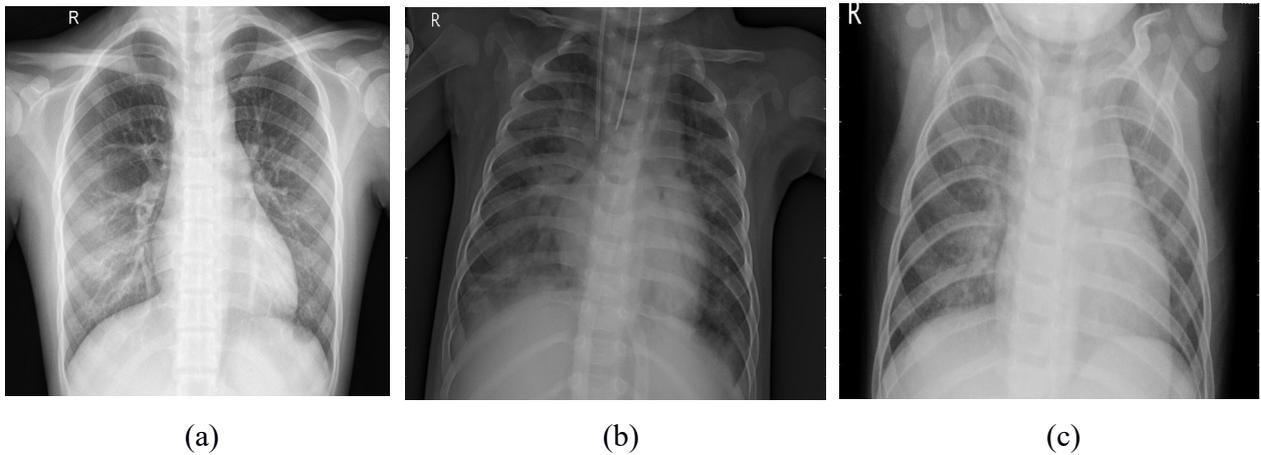

(a)            (b)            (c)

Fig 4. Chest X-Ray images: (a) Normal, (b) pneumonia bacterial and (c) pneumonia viral [48]

In structure of the DNN_NP, feature extractor has six convolution layers and single dense layer. The size of the filters was set to 3x3 for all the six layers; and the number of feature planes was 90, 90, 90, 90, 90 and 90, respectively. The single dense layer has 4 output nodes and 90x5x5x4 parameters. In structure of the DNN_VB, feature extractor has six convolution layers and single dense layer. The size of the filters was set to 3x3 for all the six layers; and the number of feature planes was 90, 90, 90, 90, 90 and 90, respectively. The single dense layer has 4 output nodes and 150x5x5x4 parameters.

Table 6 shows the classification results obtained from the studies [50-55] using other databases. Table 7 shows the classification results obtained by using the DNN_NP, DNN_VB and Kermany et al. [64]. Table 8 shows the classification results obtained from the studies [48, 56-64] using same database [65] with this study. In the tables, classification results, the need for augmentation processes and the need for preprocessing stages are shown.

In structure of the DNN_NVB, feature extractor has six convolution layers and single dense layer. The size of the filters was set to 3x3 for all the six layers; and the number of feature planes was 60, 60, 60, 60, 60 and 60, respectively. The single dense layer has 4 output nodes and 60x5x5x4 parameters. 90% performance was achieved for 3 classes by using DNN_NVB. It has not been observed that Chest X-Ray images were divided into 3 classes in studies using this database. It is an expected situation that the performance will decrease as the number of classes increases.

**3.4 Detection of BGA Solder Defects**

In recent years, high successes in classification of images have been obtained by using deep learning. Because the features are automatically extracted in deep learning, there is no need to use complex processes to determine the features. For a deep neural network to work efficiently in real-time applications, the size of the network, i.e., the number of layers in the network should be small. In our previous study [66], it was aimed to determine the defects in BGA X-ray images with high success rates by using a new deep neural network model with less number of layers. In the study [66], the proposed DNN model contained feature extractor layers and a minimum distance classier. Since the proposed network consisted of less number of layers (four convolution layers and one fully connected layer), determination of the hyper-parameters of the network and training of the network were accomplished in a short time.

Table 6: The classification results obtained from the studies using other databases [48].

| Studies | Binary Classification Results (Normal-Pneumonia) | Usage of Transfer learning | Network | The need for augmentation processes | The need for preprocessing stages |
|---|---|---|---|---|---|
| Li et al. [50] | 83.5% | Yes | SE-ResNet | Yes | Yes |
| Varshni et al. [51] | 80.02% | Yes | DenseNet-169 | No | No |
| Li et al. [52] | 92.79% | No | CNN+FCNN | Yes | No |
| Aledhari et al. [53] | 75% | Yes | VGG16 | Yes | No |
| Tilve et al. [54] | 96.2% | Yes | VGG16 | Yes | Yes |
| O'Quinn et al. [55] | 72% | Yes | AlexNet | No | No |

Table 7: The classification results obtained by using the DNN_NP, DNN_VB and DNN_NVB [48].

| DNN | Binary Classification Results (Normal-Pneumonia) | Binary Classification Results (Pneumonia bacterial-pneumonia viral) | Average Classification Results | The need for augmentation processes | The need for preprocessing stages |
|---|---|---|---|---|---|
| In our study [48] DNN_NP | **100%** | | 96% | No | No |
| In our study [48] DNN_VB | | 92% | | No | No |
| Kermany et al. [64] | 90.7% | **92.8%** | 91.75% | No | No |

Table 8: The classification results obtained from the studies using same database with this study [48].

| Studies | Binary Classification Results (Normal-Pneumonia) | Usage of Transfer learning | Network | The need for augmentation processes | The need for preprocessing stages |
|---|---|---|---|---|---|
| Vijendran et al. [53] | 92.5% | No | FCNN | No | No |
| Ayan et al. [46] | 87% | Yes | VGG16 | Yes | No |
| Islam et al. [54] | 97.34% | No | CNN+FCNN | No | Yes |
| Bhagat et al. [55] | 79.5% | | AlexNet | Yes | Yes |
| Mahajan et al. [56] | 88.78% | Yes | Inception V3 | No | No |
| Sharma et al. [57] | 90.68% | No | CNN+FCNN | Yes | No |
| Labhane et al. [58] | 98% | Yes | VGG16 | Yes | No |
| Talo et al. [59] | 97.4 | Yes | ResNet-152 | No | No |
| Stephen et al. [60] | 95.3% | No | CNN+FCNN | Yes | No |
| Kermany et al. [61] | 90.7% | Yes | Inception V3 | No | No |
| In our study [48] | **100%** | No | CNN+MDN | No | No |

In the study [66], BGA X-ray images were categorized into four classes according to the conditions of the solder joints: normal, short-circuit, bonding defect and void defect. The dataset used in this study was comprised of 67, 76, 53 and 76 images for these classes, respectively. 80% of 12 all data was allocated for the training set and the remaining 20% was allocated for the test set. Figure 5 shows the BGA X-ray sample images of the four classes. Table 9 shows the classification results of the proposed DNN for the BGA X-ray images. Table 10 shows the comparative performance results for the classification of the BGA X-ray images by using the other methods. In Table 10, the number of classes, the use of extra features or processing stages which are needed for the classification, and average test classification performance are shown.

**Table 9.** Classification results of the proposed DNN for BGAX-ray images [66].

| Class name | Training accuracy (%) | Test accuracy (%) |
|---|---|---|
| Normal | 100 | 88 |
| Short circuit | 100 | 100 |
| Bonding defect | 100 | 100 |
| Void defect | 100 | 100 |

**3.5 Classification patterns of benchmark datasets**

To demonstrate the capabilities of DivNet, it has been tested on four well-known databases in the literature, namely, the CIFAR10, CIFAR20, MNIST and IRIS databases. Table 11 summarizes the average test classification performances, the number of weights used in the networks and presence of augmentation process.

**Table 10.** Comparison results for classification of BGAX-ray images by using the other methods [66].

| Methods | Number of classes | Complexity: Feature extraction or Preprocessing | Average test classification accuracy (%) |
| --- | --- | --- | --- |
| Roh et al. [67] | 3 | morphological processes | 90 |
| Peng et al. [68] | 2 | morphological processes | 93.4 |
| Sankaran et al. [69] | 3 | removing the blurring on images | 75 |
| In our study [66] proposed DNN | 4 | no extra features or processing stages | 97 |

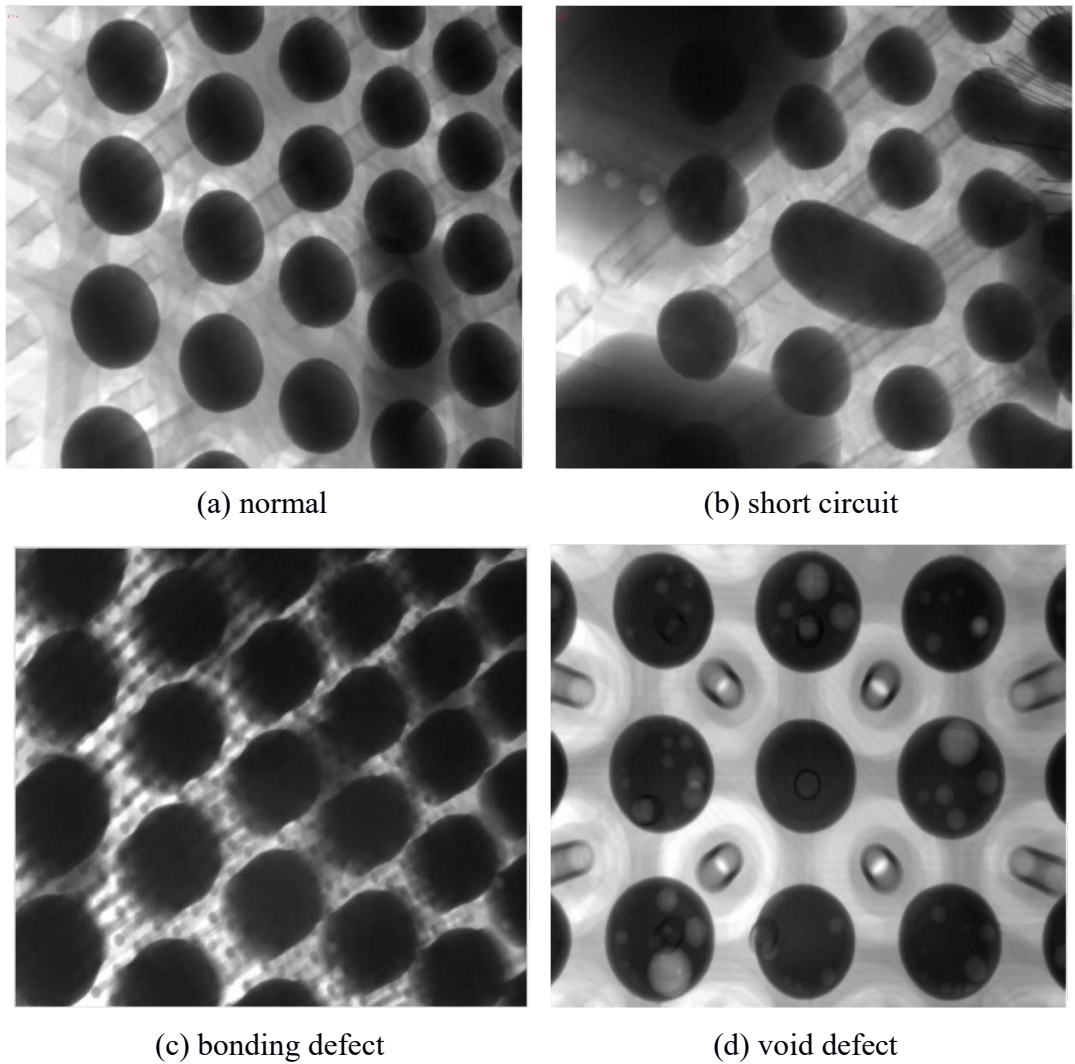

(a) normal  (b) short circuit

(c) bonding defect  (d) void defect

**Figure 5.** BGA X-ray sample images showing the four different solder joint conditions; (a) normal, (b) short circuit, (c) bonding defect, and (d) void defect [66].

The DivNet$_{MNIST}$ used for the analysis of 2D signals has five convolution layers. At the output of each convolution layer, there were batch normalization process and ReLU layer. The size of input signal was 28×28 pixels. The sizes of filters were 3×3, 5×5, 7×7, 9×9 and 28×28 for all the first, second, third, fourth and fifth convolution layers, respectively; and the numbers of feature planes were 20, 20, 20, 20 and 10 for all these layers, respectively. 86% of all data was allocated for the training set (60000 images) and the remaining 14% was allocated for the test set (10000 images). For the MNIST database, the best performance of 99.79% was obtained in a study by Wan et al [70]. The training set was augmented by rotating, scaling and cropping processes. In that study [70], a new method was also proposed to remove the unnecessary nodes from the network during the training.

The DivNet$_{CIFAR10}$ used for the analysis of 2D signals has nine convolution layers. At the output of each convolution layer, there were batch normalization process and ReLU layer. The size of the input signal was 32×32 pixels. The sizes of filters were 3×3, 5×5, 7×7, 9×9, 11×11, 13×13, 15×15, 17×17 and 32×32 for the first, second, third, fourth, fifth, sixth, seventh, eighth and ninth convolution layers, respectively; and the numbers of feature planes were 150, 150, 150, 150, 150, 150, 150, 150 and 32 for these layers, respectively. Max-pooling layers were not used. 83% of all data was allocated for the training set (50000 images) and 17% of data was allocated for the test set (10000 images). For the CIFAR10 database, the best performance of 95.5% was obtained in a study by Graham et al [71]. In the study [71], it is observed that CIFAR10 dataset has been augmented and their proposed network did not have a fully connected layer, but the network was extremely large such as the Vgg16 and AlexNet. In our study [8], the CIFAR10 dataset was not augmented, the training was carried out in one step and similar classification performances were achieved by using a smaller network structure.

The DivNet$_{CIFAR20}$ used for the analysis of 2D signals has nine convolution layers. At the output of each convolution layer, there were batch normalization process and ReLU layer. The size of the input signal was 32×32 pixels. The sizes of filters are 3×3, 3×3, 3×3, 5×5, 5×5, 5×5, 7×7, 7×7, 7×7, 9×9 and 32×32 for the first, second, third, fourth, fifth, sixth, seventh, eighth, ninth, eleven and twelve convolution layers, respectively; and the numbers of feature planes were 200, 200, 200, 200, 200, 200, 200, 200, 200, 200 and 32 for these layers, respectively. Max-pooling layers were not used. 83% of all data was allocated for the training set (50000 images) and 17% of data was allocated for the test set (10000 images). For the CIFAR20 database, the best performance of 69.1% was obtained by using ResNet-32 in a study by Cui et al [72]. In the study [72], it is observed that CIFAR20 dataset has been augmented, and complex methods have been tried to obtain highest performance. In our study [8], the CIFAR20 dataset was not augmented, and similar classification performances were achieved by using a smaller network structure.

**Table 11** Performance results for classification of IRIS, MNIST, CIFAR10 and CIFAR20 databases [8]

| Methods | Performances Results | | | | | |
|---|---|---|---|---|---|---|
| | Test set acc. for MNIST (%) | Test set acc. for CIFAR10 (%) | Test set acc. for CIFAR20 (%) | Test set acc. for IRIS (%) | Number of weights in the network | Augmentation process |
| Wan et al [71] (only MNIST) | 99.79 | | | | 4886420 | Yes |
| DivNet in [8] (only MNIST) | 99.8 | | | | 171780 | No |
| Graham [70] (only CIFAR10) | | 95.5 (for single test) | | | 184174800 | Yes |
| DivNet in [8] (only CIFAR10) | | 91 | | | 24039150 | No |
| Cui [72] (only CIFAR20) | | | 69.1 | | 21051072 | Yes |
| DivNet in [8] (only CIFAR20) | | | 69 | | 19399000 | No |
| DivNet in [8] (only IRIS) | | | | 100 | 820 | No |

Even when the size of input signal is small, proposed network can also be applied into classification problems. The DivNet$_{IRIS}$ used for the analysis of 1D signals has 4 convolution layers [8]. At the output of each convolution, layer there was a ReLU layer. The size of input signal was 4 (there are four features). The sizes of filters were 2, 2, 2, and 4 for all the first, second, third and four convolution layers, respectively; and the numbers of feature planes were 10, 10, 10 and 10 for all these layers, respectively. Max-pooling layers were not used. 80% of all data was allocated for the training set (120 vectors), and the remaining 20% of data was allocated for the test set (30 vectors). The performance of 100% was obtained for training and test set. Number of weights (not node) in the network is 820.

**3.6 Detection of the Murmurs in the Heart Sound**

In this study, for the first time, sound records are classified as background sound (voice), normal heart sound and murmur by using the DivFE. And at the same time, the DivFE will be tested as an application running in real time. A new set is created using three different heart sound datasets (Pascal, PhysioNet, and Kaggle datasets) and background sound. The sampling frequency of the heart sounds is different in each dataset. In new heart sound dataset, an arrangement is made so that the length of each recording is 3 seconds, and the sampling frequency of the recordings is 1000 Hz. New set contains 3×140 records with background sound, normal and murmur heart sounds. 80% of all data is allocated for the training set and the remaining 20% is allocated for the test set. In this study, the augmentation process introduced in [17] is applied to the heart sounds. After augmentation process, the training set is increased three times with the original heart sound also included in the augmented set. In fact, for $N$ randomly generated training and test sets, the DivFE should be trained $N$ times and the average of $N$ accuracies for the test sets should be calculated to achieve statistically significant success. In this study, the classification result of only 1 trial ($N=1$) is given.

The $DivFE_{HeartSound}$ used for the analysis of heart sound has eight convolution layers. At the output of each convolution layer, there are batch normalization process and ReLU layer. The batch normalization and drop-out functions are used between each layer to further strengthen the training. The size of input signal is 3000×1 pixels. The sizes of filters are 192, 96, 48, 24, 12, 6, 3 and 24 for all the first, second, third, fourth, fifth, sixth, seventh and eight convolution layers, respectively; and the numbers of feature planes are 40, 40, 40, 40, 40, 40, 40 and 16 for all these layers, respectively. For new dataset, the best performance (100%) according to the literature is obtained for three classes by using the $DivFE_{HeartSound}$. Table 12 shows the performance results obtained by using DNNs.

**Table 12** Performance results for classification of the Heart sounds by using DNNs.

| Methods | Number of classes | Accuracy % | Database |
|---|---|---|---|
| Dominguiz-Morales et al. [73] | 2 | 97 | PhysioNet |
| Mishra et al. [74] | 2 | 95 | PhysioNet |
| Khan et al.[76] | 2 | 95.4 | PhysioNet |
| Demir et al. [75] | 2 | 80 | Pascal |
| Khan et al.[76] | 2 | 96.8 | Pascal |
| The proposed study | 3 | 100 | Pascal, PhysioNet, and Kaggle |

In this context, an application is developed on mobile phone for test purposes. Detailed information about the development environment of the program is given below:

In this study, we have developed a system to classify heart sounds by running DivFE on an Android mobile phone (Samsung Note 8). To realize the system software, Ubuntu 18.04 (Linux), Android Studio and the Qt5 IDE are installed; Android and desktop kits are created on Qt5 IDE. The cmake software is installed to compile the OpenCV source code. OpenCV is loaded from internet and compiled with cmake for both Android and Linux desktop kits. The compiled OpenCV is made compatible with Qt5 to access the OpenCV library on a mobile phone. Then, a program in C++ is coded to get the heart sounds from the audio input of the mobile phone in real-time. The DivFE is trained on a different workstation with GPU cards by using benchmark datasets and compressed into a "*.bp" form to be loaded with OpenCV. The DivFE is enclosed in an apk file to bypass security problems. The OpenCV libraries are loaded on the mobile phone by Qt5 audio application, and DivFE is loaded by using OpenCV. The hardware (shown in Fig. 6) for the application is very simple and cheap (< 5 USD). The microphone of the headset is disconnected, and then, mechanically connected to the stethoscope. The hardware is also suitable to acquire heart sound data from the audio input by using the MATLAB or Python. There is also a Maass-Weber filter in the audio program that runs in real time. While the audio application plots the heart sounds on the screen, it also transmits the sound signal to the phone's headset, and records the heart sounds as wav file in the download directory of the phone. In the meantime, DivFE analyses heart sounds at time intervals of 3 seconds in real time and determines whether there is any murmur in the audio. In addition, S1 sounds are detected in the audio recordings to determine the RR intervals (heartbeat intervals). RR intervals are analyzed to determine whether there is an arrhythmia (abnormal heartbeat) or not.

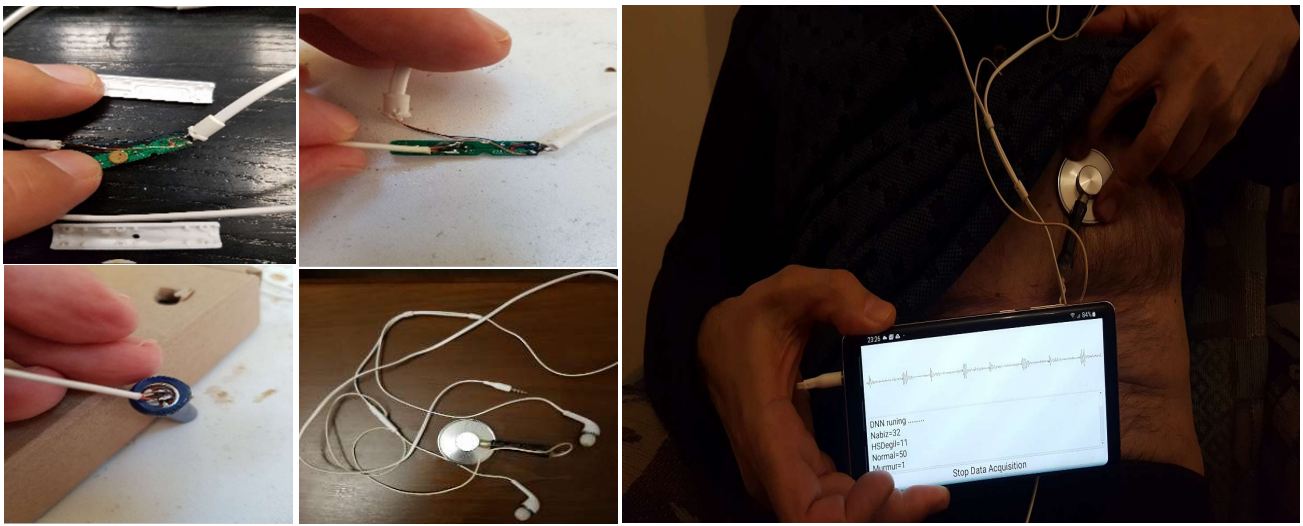

(a) Hardware                                                                                   (b) Application

Figure 6. An application of DivFE analyzing heart sounds in real-time.

The audio application running on mobile phone is controlled by USB debugging in real time. DivFE's speed, performance and memory needs (the size of DivFE is 459600 Bytes) are tested in real time for heart sounds. It is not possible to install and run large-sized DNN structures, such as the AlexNet, on such portable systems. By this project, we had the chance to observe the speed, memory requirement, and success of DivFE in a real time application. We have seen how useful it is.

## 4. Discussions and Conclusions

In this study, the performance of DivFE in other areas, including biomedical signals and images, is investigated. In different areas, it has been observed [8, 17, 48, 66] that a higher classification performance was obtained by using DivFE with less number of nodes.

DNN structures are continuously developing and achieving high performances in classification problems. Also, it is observed that success rates obtained with DNNs are higher than those obtained with traditional neural networks. In addition, one of the advantages of DNNs is that there is no need to spend an extra effort to determine the features; the CNN automatically extracts the features from the dataset during the training. Besides their benefits, the DNNs have the following three major drawbacks among the others: (1) Researchers have struggled with over-fitting and under-fitting issues in the training of DNNs, (2) determination of even a coarse structure for the DNN may take days, and (3) most of the time, the DNN structures were too large to be too bulky to be used in real time applications.

In previous studies [8, 17, 48, 66], we have addressed especially these disadvantages. A novel training and structure for deep neural network were introduced. We have modified the training and structure of DNN to increase the classification performance, to determine a coarse structure of DNNs easily, and to be used with less number of nodes in DNN structure. A minimum distance network (MDN) following the last layer of the convolutional neural network (CNN) was used as the classifier instead of a fully connected neural network (FCNN). The weights of the nodes of the MDN have not changed during the training. Therefore, the proposed method focuses solely on the training of the CNN (i), as the minimum distance network does not require any training. After the training is complete, we can compute the divergence analysis using the vectors at output of the CNN. This analysis will show us how valuable features CNNs find. For any set as input, the distribution of vectors at the output of the DivFE is expected to bring Eq. 2 to the highest value. In this case, the DivFE will have the best features. In this regard, there needs to locate the mean vectors (centers) of the classes at the farthest positions in the feature space with equal distances from each other. The Walsh vectors showed up to meet these requirements. Hence, during the training, the CNN part of the DivFE is only trained to map the input vectors to Walsh vectors (ii). In this case, according to Eq. (2), the ratio of a larger between-class scatters (ii) to a smaller within-class scatters (i) leads to a high divergence value to be obtained. This approach leads us to better training of the CNN in the DivFE.

MDN following the flatten layer of the convolutional neural network (CNN) is used as the classifier instead of a fully connected neural network (FCNN). Since the proposed network consists of less number of layers, determination of the hyper-parameters of the network is accomplished in a short time. The smaller the DNN structure, the easier it is to search for a coarse model.

In this study, the proposed DNN is comprised of a CNN and a minimum distance network as the classifier instead of the FCNN. In this study, with the developed training strategy, it is possible to train the CNN and classifier (such as fully connected layers) individually. If both are desired to be included in the network, first the convolutional layer and then the classifier can be trained sequentially. If they are trained separately, the features are more likely to be accurately determined. As a result of the achievements in the extracted features, the fully connected neural layers can be eliminated. Therefore, we preferred to employ only a simple minimum distance classifier instead of a fully connected layer. Thus, it may be possible to classify signals or images by using fewer nodes.

The problem of over-fitting is frequently encountered in training of DNNs. The augmentation process must be not only preferred to balance the number of data in each class, and also should be used to eliminate the over-fitting problem. For this reason, augmentation processes should be applied to signal's some parameters which do not contain information. In our previous study [17], amplification of signal by random amount, polarity inversion of signal, rotation of signal along time dimension, and random noise injection to signal were preferred as augmentation processes. In [17], we have investigated the effect of the augmentation process on the classification performance of MI EEG signals. This augmentation process is also used in the classification of the heart sounds. In these studies, discriminative information is in the frequency spectrum of signals. During training, these augmentation operations prevents the DNN from use of a wrong information (such as amplitude values in time-domain), and direct the DNN to examine the spectrum of the signal with filters.

The DivFE consists of convolutional layer and MDN network. MDN consists of Walsh vectors (rows or columns of Walsh matrix). MDN network is not used in the DivFE's training. After the last convolutional layer, there is a flatten layer. During the training, the number of convolutional layers in the DivFE can be determined automatically. Firstly, classification success is obtained for the training set by using only 1 convolutional layer and 1 flatten layer. If the classification performance for the training set (not test set) is less than a selected value (for example 95%), the number of convolutional layer in the DivFE is increased to 2 and 1 flatten layer is used again. In this way, convolutional layers are increased until the classification success for the training set is above 95%. During this training process, the weights of the previous convolutional layer can be kept or the weights of all convolutional layers can be modified again. Since DivFE does not have a fully connected neural network, it is quick and easy to determine the number of layers of CNN.

**Acknowledgement**

These works in the article were supported by the Istanbul Technical University Scientific Research Project Unit [ITU-BAP MYL-2018-41621 and ITU-BAP MYL-2019-41895].